\documentclass[11pt]{article}

\usepackage{ACL2023}

\usepackage{times}
\usepackage{latexsym}

\usepackage[T1]{fontenc}

\usepackage[utf8]{inputenc}

\usepackage{microtype}
\usepackage{multirow}
\usepackage{graphicx}
\usepackage{enumitem}
\usepackage{fancyhdr}
\usepackage{inconsolata}
\usepackage{pgfplots}
\pgfplotsset{compat=1.17}
\newcommand\blfootnote[1]{%
  \begingroup
  \renewcommand\thefootnote{}\footnote{#1}%
  \addtocounter{footnote}{-1}%
  \endgroup
}

\title{OffMix-3L: A Novel Code-Mixed Dataset in Bangla-English-Hindi for Offensive Language Identification}

\author{Dhiman Goswami\textsuperscript{*}, Md Nishat Raihan\textsuperscript{*}, Antara Mahmud \\  \textbf{Antonios Anastasopoulos}, \textbf{Marcos Zampieri} \\
        George Mason University, USA \\
        \texttt{\{dgoswam, mraihan2, amahmud4, antonis, mzampier\}@gmu.edu} \\
        }

\begin{document}
\maketitle
\begin{abstract}
Code-mixing is a well-studied linguistic phenomenon when two or more languages are mixed in text or speech. Several works have been conducted on building datasets and performing downstream NLP tasks on code-mixed data. Although it is not uncommon to observe code-mixing of three or more languages, most available datasets in this domain contain code-mixed data from only two languages. In this paper, we introduce OffMix-3L, a novel offensive language identification dataset containing code-mixed data from three different languages. We experiment with several models on this dataset and observe that BanglishBERT outperforms other transformer-based models and GPT-3.5. \blfootnote{*These two authors contributed equally to this work.} \blfootnote{\bf WARNING: This paper contains examples that are offensive in nature.}
\end{abstract}

\section{Introduction}

Code-mixing and code-switching are common linguistic phenomena observed both in speech and text form. While the two terms are often used interchangeably, code-mixing is defined as the use of words or morphemes from multiple languages within a single utterance, sentence, or discourse whereas code-switching refers to the deliberate alternation between multiple languages within the same context \cite{thara2018code}. The first case is often spontaneous while the second case is purposeful. However, both are widely observed in bilingual and multilingual communities. 

As described in \citet{anastassiou2017factors}, several social, linguistic, and cognitive factors are behind these two phenomena. Socially, this often serves as a sign of group identity which allows individuals to navigate multiple social and cultural affiliations. Linguistically, it is common for bilingual speakers to not be able to find a word for a specific concept in one language thus using a word from another language to help communication. Additionally, there are several cases even in monolingual communities, when code-mixing might be the convenient way to express a concept as in the case of English loan words such as {\em feedback} used in various languages.  

Most commonly, code-mixing is a bilingual phenomenon. \citet{byers2013bilingualism}, for example, estimates that by the year 2035, over half of children enrolled in kindergarten in California will have grown up speaking a language other than English. Another study conducted by \citet{jeffery2020language} shows that bilingualism is a common practice in European countries such as Germany and Spain. However, in cosmopolitan cities and areas like New York, London, Singapore, and others, code-mixing with three or more languages is fairly common. This is also observed in countries like Luxembourg, and regions such as West Bengal, and South-East India where more than two languages are commonly used on a daily basis.

Several papers have presented code-mixed datasets for various NLP tasks \cite{khudabukhsh2020harnessing,krishnan2022cross}. However, most of these datasets are bilingual leaving the processing of code-mixing in three or more languages largely unexplored. In this paper, we present a Bangla-Hindi-English dataset annotated for offensive language identification. To the best of our knowledge, this is one of the first datasets to contain code-mixing between more than two languages.  

The main contributions of this paper are as follows:

\begin{itemize}
    \item We introduce OffMix-3L, a novel three-language code-mixed test dataset in Bangla-Hindi-English for offensive language identification. OffMix-3L contains 1,001 instances annotated by speakers of the three languages. We made OffMix-3L freely available to the community.\footnote{https://github.com/LanguageTechnologyLab/OffMix-3L}
    \item We provide a comprehensive evaluation of several monolingual, bilingual, and multilingual models on OffMix-3L.
\end{itemize}

\noindent We present OffMix-3L exclusively as a test set due to the unique and specialized nature of the task. The size of the dataset, while limiting for training purposes, offers a high-quality testing environment with gold-standard labels that will serve as a benchmark in this domain. Given the scarcity of similar datasets and the challenges associated with data collection, OffMix-3L provides an important resource for the rigorous evaluation of offensive language identification models, filling a critical gap in multi-level code-mixing research. 


\section{Related Work}

There have been several studies describing Bangla-English \cite{alam2006code, hasan2015reviewing, hossain2015case, begum2013code, mahbub2016sociolinguistic}, Hindi-English \cite{singh1985grammatical, bali2014borrowing, thara2018code} and Bangla-Hindi \cite{ali2019effects, jose2020survey} code-mixing and code-switching separately. Code-mixing between these three languages has also been studied separately in NLP. 

There have been few studies conducted on offensive language identification for Bangla-English code-mixed data. The work by \citet{jahan2019abusive} focused on detecting Bangla-English code-mixed and transliterated offensive comments on Facebook. Another Bangla-English dataset is gathered by \citet{mridha2021boost}, where they collected 2,200 instances. 

Comparatively more work has been carried out for Hindi-English Code-mixing. \citet{sreelakshmi2020detection} uses fastText \cite{DBLP:journals/corr/JoulinGBDJM16} to represent 10,000 instances collected from different sources. Other offensive language datasets collected from Facebook and Twitter were introduced by \citet{bohra2018dataset, KUMAR18.861, rani2020comparative}. \citet{mundra2022fa} proposes Fused Attention-based Network (FA-Net), which introduces a fusion of attention mechanism of collective and mutual learning between local and sequential features for Hindi-English offensive language and hate speech classification. \citet{gupta2021hindi} uses Character Level Embeddings, GRU, and attention layer to offensive language identification in Hindi-English code-mixed.

To the best of our knowledge, no existing work focuses specifically on Hindi-Bangla code-mixing. However, some studies focused on multiple Indian languages code-mixing altogether including Bangla and Hindi. The work by \citet{vasantharajan2021hypers} focuses on offensive language identification in Dravidian languages. A few similar works include \citet{ravikiran2021dosa,sai2020siva,kumar2020evaluating, kumari2020ai_ml_nit_patna,ranasinghe2021evaluation}.

In summary, to the best of our knowledge, there has been no work on offensive language identification for code-mixed Bangla-English-Hindi. There have also been no offensive language datasets made available for these three languages. OffMix-3L fills this gap by providing the community with a novel resource for these three languages. OffMix-3L provides the community with the opportunity to evaluate how state-of-the-art models perform on Bangla-English-Hindi. 

\section{The \textit{OffMix-3L} Dataset}

We choose a controlled data collection method, asking the volunteers to freely contribute data in Bangla, English, and Hindi. This decision stems from several challenges of extracting such specific code-mixed data from social media and other online platforms. Our approach ensures data quality and sidesteps the ethical concerns associated with using publicly available online data. Such types of datasets are often used when it is difficult to mine them from existing corpora. As examples, for fine-tuning LLMs on instructions and conversations, semi-natural datasets like \citet{dolly2023} and \citet{awesome_instruction_datasets} have become popular. 

\paragraph{Data Collection} 
A group of 10 undergraduate students fluent in the three languages was asked to prepare 250 to 300 social media posts each. They were allowed to use any language including Bangla, English, and Hindi to prepare posts on several daily topics like politics, sports, education, social media rumors, etc. We also ask them to switch languages if and wherever they feel comfortable doing so. The inclusion of emojis, hashtags, and transliteration was also encouraged. The students had the flexibility to prepare the data as naturally as possible. Upon completion of this stage, we gathered 1,734 samples that contained at least one word or sub-word from each of the three languages using langdetect \cite{langdetect} an open-sourced Python tool for language identification.  

\paragraph{Data Annotation} We annotate the dataset in two steps. Firstly, we recruited three students from social science, computer science, and linguistics fluent in the three languages to serve as annotators. They annotated all 1,734 samples with one of the two labels (Non-Offensive and Offensive) with a raw agreement of 63.7\%. We then take 1,106 instances, where all three annotators agree on the labels, and use them in a second step. To further ensure high-quality annotation, we recruit a second group of annotators consisting of two NLP researchers fluent in the three languages. After their annotation, we calculate a raw agreement of 91\% \cite{kvaalseth1989note}, a Cohen Kappa score of 0.82. After the two stages, we only keep the instances where both annotators agree, and we end up with a total of 1,001 instances. The label distribution is shown in Table \ref{tab:label_distribution}. 

\begin{table} [!h]
\centering
\scalebox{.92}{
\begin{tabular}{lcc}
\hline
\textbf{Label} & \textbf{No. of Data} & \textbf{Percentage}\\
\hline
Non-Offensive & 522 & 52.15\% \\
Offensive & 479 & 47.85\% \\

\hline
Total & 1,001 & 100\% \\
\hline
\end{tabular}
}
\caption{Label distribution in OffMix-3L}
\label{tab:label_distribution}
\end{table}

\paragraph{Dataset Statistics} A detailed description of the dataset statistics is provided in Table \ref{tab:data_card}. Since the dataset was generated by people whose first language is Bangla, we observe that the majority of tokens in the dataset are in Bangla. There are several \textit{Other} tokens in the dataset that are not from Bangla, English, or Hindi language. The \textit{Other} tokens in the dataset primarily contain transliterated words as well as emojis and hashtags. Also, there are several misspelled words that have been classified as \textit{Other} tokens too. 

\begin{table} [!h]
\centering
\scalebox{.82}{
\begin{tabular}{l|c|cccc}
\hline
 & \textbf{All} & \textbf{Bangla} & \textbf{English} & \textbf{Hindi} & \textbf{Other}\\
\hline
Tokens & 87,190 & 31,228 & 6,690 & 14,694 & 34,578\\
Types & 18,787 & 7,714 & 1,135 & 1,413 & 8,645\\
Avg & 87.10 & 31.20 & 6.68 & 14.68 & 34.54\\
Std Dev & 20.58 & 8.60 & 3.05 & 5.74 & 10.98\\
\hline
\end{tabular}
}
\caption{OffMix-3L Data Card. The row {\em Avg} represents the average number of tokens with its standard deviation in row {\em Std Dev}.}
\label{tab:data_card}
\end{table}

\paragraph{Synthetic Train and Development Set} We present OffMix-3L as a test dataset and we build a synthetic train and development set that contains Code-mixing for Bangla, English, and Hindi. We use two English training datasets annotated with the same labels as OffMix-3L, namely OLID \cite{zampieri-etal-2019-semeval} and SOLID \cite{rosenthal-etal-2021-solid}. We randomly select 100,000 data instances randomly and we carefully choose an equal number of Non-Offensive and Offensive instances. We then use the \textit{Random Code-mixing Algorithm} \cite{krishnan2022cross} and \textit{r-CM}  \cite{santy2021bertologicomix} to generate the synthetic Code-mixed dataset. 

\section{Experiments}


\paragraph{Monolingual Models} We use seven monolingual models for these experiments, five general models, and two task fined-tuned ones. The five monolingual models are DistilBERT \cite{DBLP:journals/corr/abs-1910-01108}, BERT \cite{devlin2019bert}, BanglaBERT \cite{kowsher2022bangla}, roBERTa \cite{liu2019roberta}, HindiBERT \cite{nick_doiron_2023}. BanglaBERT is trained in only Bangla and HindiBERT on only Hindi while DistilBERT, BERT, and roBERTa are trained in English only. Finally, the two English task fine-tuned models we use are HateBERT and fBERT \cite{caselli-etal-2021-hatebert, sarkar-etal-2021-fbert-neural}.

\paragraph{Bilingual Models} BanglishBERT \cite{bhattacharjee-etal-2022-banglabert} and HingBERT \cite{nayak-joshi-2022-l3cube} are used as bilingual models as they are trained on both Bangla-English and Hindi-English respectively.

\paragraph{Multilingual Models} We use mBERT \cite{devlin2019bert} and XLM-R \cite{conneau2019unsupervised} as multilingual models which are respectively trained on 104 and 100 languages including Bangla-English-Hindi. We also use IndicBERT \cite{kakwani2020indicnlpsuite} and MuRIL \cite{khanuja2021muril} which cover 12 and 17 Indian languages, respectively, including Bangla-English-Hindi. We also perform hyper-parameter tuning while using all the models to prevent overfitting.

\paragraph{Prompting} We use prompting with GPT-3.5-turbo model \cite{openai2023gpt35turbo} from OpenAI for this task. We use the API for zero-shot prompting (see Figure \ref{fig:prompt1}) and ask the model to label the test set.\\

Additionally, we run the same experiments separately on synthetic and natural datasets splitting both in a 60-20-20 way for training, evaluating, and testing purposes.

\begin{figure}[h]
\centering
\scalebox{.92}{
\begin{tikzpicture}[node distance=1cm]
    \tikzstyle{block} = [rectangle, draw, fill=blue!20, text width=\linewidth, text centered, rounded corners, minimum height=4em]
    \tikzstyle{operation} = [text centered, minimum height=1em]
    \node [block] (rect1) {\textbf{Role:}{ "You are a helpful AI assistant. You are given the task of offensive text classification. }};
    \node [operation, below of=rect1] (plus1) {};
    \node [block, below of=plus1] (rect2) {\textbf{Definition:}{ An offensive text is something that upsets or embarrasses people because it is rude or insulting. You will be given a text to label either 'Offensive' or 'Non-Offensive'. }};
    \node [operation, below of=rect2] (plus2) {};
    \node [block, below of=plus2] (rect3) {\textbf{Task:}{ Generate the label for this \textbf{"text"} in the following format: \textit{<label> Your\_Predicted\_Label <$\backslash$label>}. Thanks."}};
\end{tikzpicture}
}
\caption{Sample GPT-3.5 prompt.}
\label{fig:prompt1}
\end{figure}

\section{Results}

In this experiment, synthetic data is used as a training set and natural data is used as the test set. The F1 scores of monolingual models range from 0.43 to 0.66 where BERT performs the best. mBERT is the best of all the multilingual models with an F1 score of 0.63. Besides, a zero-shot prompting technique on GPT 3.5 turbo provides a 0.57 weighted F1 score. The best task fine-tuned model is HateBERT with the F1 score of 0.60. Among all the models BanglishBERT scores 0.68 which is the best achieved F1 score. These results are available in Table \ref{Results1}. 

\begin{table} [!h]
\centering
\scalebox{0.92}{
\begin{tabular}{lcc}
\hline
\textbf{Models}  & \textbf{F1 Score} \\
\hline
BanglishBERT & \textbf{0.68} \\
BERT & 0.66  \\
mBERT & 0.63 \\
HingBERT & 0.60  \\
MuRIL & 0.60 \\
HateBERT & 0.60 \\
fBERT & 0.58 \\
roBERTa & 0.58  \\
XLM-R & 0.57 \\
DistilBERT & 0.57  \\
GPT 3.5 Turbo & 0.57 \\
BanglaBERT & 0.54 \\
IndicBERT & 0.55 \\
HindiBERT & 0.43  \\
\hline
\end{tabular}
}
\caption{Weighted F-1 score for different models: training on synthetic and tested on natural data (OffMix-3L).}
\label{Results1}
\end{table}

\noindent We perform the same experiment using synthetic data for training and testing. We present results in Table \ref{tab_synthetic}. Here, mBERT and XLM-R with 0.88 F1 scores are the best-performing models. 


\begin{table} [!h]
\centering
\scalebox{.92}{
\begin{tabular}{lcc}
\hline
\textbf{Models}  & \textbf{Weighted F1 Score} \\
\hline
XLM-R & \textbf{0.88} \\
mBERT & \textbf{0.88} \\
BanglishBERT & 0.86 \\
BERT & 0.83  \\
HingBERT & 0.82  \\
IndicBERT & 0.82 \\
MuRIL & 0.81 \\
fBERT & 0.81 \\
HateBERT & 0.81 \\
roBERTa & 0.80  \\
DistilBERT & 0.79  \\
BanglaBERT & 0.76 \\
HindiBERT & 0.73  \\
\hline
\end{tabular}
}
\caption{Weighted F-1 score for different models: training on synthetic and tested on synthetic data.}
\label{tab_synthetic}
\end{table}

\section{Error Analysis}

We observe \textit{Other} tokens in almost 39\% of the whole dataset, as shown in Table \ref{tab:data_card}. These tokens occur due to transliteration which poses a challenge for most of the models since not all of the models are pre-trained on transliterated tokens. BanglishBERT did well since it recognizes both Bangla and English tokens. However, the total number of tokens for Hindi-English is less than Bangla-English tokens, justifying HingBERT's inferior performance compared to BanglishBERT (see Table \ref{Results1}). Also, misspelled words and typos are also observed in the datasets, which are, for the most part, unknown tokens for the models, making the task even more difficult. Some examples are available in Appendix \ref{sec:appendix_a} which are classified wrongly by all the models.

\section{Conclusion and Future Work}

In this paper, we presented OffMix-3L, a Bangla-English-Hindi code-mixed offensive language identification dataset containing 1,001 instances. We also created 100,000 synthetic data in the same three languages for training. We evaluated various monolingual models on these two datasets. Our results show that when training on synthetic data and testing on OffMix-3L, BanglishBERT performs the best. When using synthetic data for both training and testing, multilingual models such as mBERT and XLM-R perform well. 
In the future, we would like to expand OffMix-3L so that it can serve as both training and testing data. Additionally, we are working on pre-training Bangla-English-Hindi trilingual code-mixing models for offensive language identification. 


\section*{Acknowledgments}

We thank the annotators who helped us with the annotation of OffMix-3L. We further thank the anonymous workshop reviewers for their insightful feedback. 
Antonios Anastasopoulos is generously supported by NSF award IIS-2125466.

\section*{Limitations}
Although most datasets for the downstream tasks are scraped from social media posts in the real world, in our case these data instances are generated in a semi-natural manner, meaning that they were generated by people but not scraped from social media directly. This was done due to the complexity of extracting contents that contain a specific 3 language code-mixing in them. Also, the dataset is comparatively smaller in size, since it is costly to generate data by a specific set of people who are fluent in all 3 target languages.

\section*{Ethics Statement}
The dataset introduced in this paper, which centers on the analysis of offensive language in Bangla-English-Hindi code-mixed text, adheres to the  \href{https://www.aclweb.org/portal/content/acl-code-ethics}{ACL Ethics Policy} and seeks to make a valuable contribution to the realm of online safety. OffMix-3L serves as an important resource for the moderation of online content, which will make it easier to create safer digital environments. Moreover, the contributors and annotators of the dataset are paid respectable remuneration and also attended two sessions with psychologist before starting and after completing the work to ensure their mental health is not compromised throughout the course of this dataset preparation.



\bibliography{custom}

\begin{thebibliography}{50}
\expandafter\ifx\csname natexlab\endcsname\relax\def\natexlab#1{#1}\fi

\bibitem[{Alam(2006)}]{alam2006code}
Suraiya Alam. 2006.
\newblock Code-mixing in bangladesh: A case study of non-government white-collar service holders and professionals.
\newblock \emph{Asian affairs}, 28(4):52--70.

\bibitem[{Ali et~al.(2019)Ali, Ranjha, and Jillani}]{ali2019effects}
Muhammad~Mooneeb Ali, Mazhar~Iqbal Ranjha, and Sartaj~Fakhar Jillani. 2019.
\newblock Effects of code mixing in indian film songs.
\newblock \emph{Journal of Media Studies}, 31(2).

\bibitem[{Anastassiou(2017)}]{anastassiou2017factors}
Fotini Anastassiou. 2017.
\newblock Factors associated with the code mixing and code-switching of multilingual children: An overview.
\newblock \emph{International Journal of Linguistics, Literature and Culture}, 4(3):13--26.

\bibitem[{Bali et~al.(2014)Bali, Sharma, Choudhury, and Vyas}]{bali2014borrowing}
Kalika Bali, Jatin Sharma, Monojit Choudhury, and Yogarshi Vyas. 2014.
\newblock “i am borrowing ya mixing?" an analysis of english-hindi code mixing in facebook.
\newblock In \emph{Proceedings of CodeSwitch}.

\bibitem[{Begum and Haque(2013)}]{begum2013code}
Most~Tasnim Begum and Md~Mahmudul Haque. 2013.
\newblock Code mixing in the ksa: A case study of expatriate bangladeshi and indian esl teachers.
\newblock \emph{Arab World English Journal}, 4(4).

\bibitem[{Bhattacharjee et~al.(2022)Bhattacharjee, Hasan, Ahmad, Mubasshir, Islam, Iqbal, Rahman, and Shahriyar}]{bhattacharjee-etal-2022-banglabert}
Abhik Bhattacharjee, Tahmid Hasan, Wasi Ahmad, Kazi~Samin Mubasshir, Md~Saiful Islam, Anindya Iqbal, M.~Sohel Rahman, and Rifat Shahriyar. 2022.
\newblock {B}angla{BERT}: Language model pretraining and benchmarks for low-resource language understanding evaluation in {B}angla.
\newblock In \emph{Findings of the ACL}.

\bibitem[{Bohra et~al.(2018)Bohra, Vijay, Singh, Akhtar, and Shrivastava}]{bohra2018dataset}
Aditya Bohra, Deepanshu Vijay, Vinay Singh, Syed~Sarfaraz Akhtar, and Manish Shrivastava. 2018.
\newblock A dataset of hindi-english code-mixed social media text for hate speech detection.
\newblock In \emph{Proceedings of PEOPLES}.

\bibitem[{Byers-Heinlein and Lew-Williams(2013)}]{byers2013bilingualism}
Krista Byers-Heinlein and Casey Lew-Williams. 2013.
\newblock Bilingualism in the early years: What the science says.
\newblock \emph{LEARNing landscapes}, 7(1):95.

\bibitem[{Caselli et~al.(2021)Caselli, Basile, Mitrovi{\'c}, and Granitzer}]{caselli-etal-2021-hatebert}
Tommaso Caselli, Valerio Basile, Jelena Mitrovi{\'c}, and Michael Granitzer. 2021.
\newblock {H}ate{BERT}: Retraining {BERT} for abusive language detection in {E}nglish.
\newblock In \emph{Proceedings of WOAH}.

\bibitem[{Conneau et~al.(2020)Conneau, Khandelwal, Goyal, Chaudhary, Wenzek, Guzm{\'a}n, Grave, Ott, Zettlemoyer, and Stoyanov}]{conneau2019unsupervised}
Alexis Conneau, Kartikay Khandelwal, Naman Goyal, Vishrav Chaudhary, Guillaume Wenzek, Francisco Guzm{\'a}n, Edouard Grave, Myle Ott, Luke Zettlemoyer, and Veselin Stoyanov. 2020.
\newblock Unsupervised cross-lingual representation learning at scale.
\newblock In \emph{Proceedings of ACL}.

\bibitem[{Databricks(2023)}]{dolly2023}
Databricks. 2023.
\newblock \href {https://www.databricks.com/blog/2023/04/12/dolly-first-open-commercially-viable-instruction-tuned-llm} {Dolly 2.0: An open source, instruction-following large language model}.
\newblock Accessed: 2023-09-10.

\bibitem[{Devlin et~al.(2019)Devlin, Chang, Lee, and Toutanova}]{devlin2019bert}
Jacob Devlin, Ming-Wei Chang, Kenton Lee, and Kristina Toutanova. 2019.
\newblock {BERT: Pre-training of Deep Bidirectional Transformers for Language Understanding}.
\newblock In \emph{Proceedings of NAACL}.

\bibitem[{Gupta et~al.(2021)Gupta, Sehra, Vardhan et~al.}]{gupta2021hindi}
Vasu Gupta, Vibhu Sehra, Yashaswi~Raj Vardhan, et~al. 2021.
\newblock Hindi-english code mixed hate speech detection using character level embeddings.
\newblock In \emph{Proceedings of ICCMC}.

\bibitem[{Hasan et~al.(2015)Hasan, Akhand et~al.}]{hasan2015reviewing}
Md~Kamrul Hasan, Mohd Akhand, et~al. 2015.
\newblock Reviewing the challenges and opportunities presented by code switching and mixing in bangla.
\newblock \emph{Journal of Education and Practice}, 6(1):103--109.

\bibitem[{Hossain and Bar(2015)}]{hossain2015case}
Didar Hossain and Kapil Bar. 2015.
\newblock A case study in code-mixing among jahangirnagar university students.
\newblock \emph{International Journal of English and Literature}, 6(7):123--139.

\bibitem[{Jahan et~al.(2019)Jahan, Ahamed, Bishwas, and Shatabda}]{jahan2019abusive}
Maliha Jahan, Istiak Ahamed, Md~Rayanuzzaman Bishwas, and Swakkhar Shatabda. 2019.
\newblock Abusive comments detection in bangla-english code-mixed and transliterated text.
\newblock In \emph{Proceedings of ICIET}.

\bibitem[{Jeffery and van Beuningen(2020)}]{jeffery2020language}
Jill~V Jeffery and Catherine van Beuningen. 2020.
\newblock Language education in the eu and the us: Paradoxes and parallels.
\newblock \emph{Prospects}, 48(3-4):175--191.

\bibitem[{Jose et~al.(2020)Jose, Chakravarthi, Suryawanshi, Sherly, and McCrae}]{jose2020survey}
Navya Jose, Bharathi~Raja Chakravarthi, Shardul Suryawanshi, Elizabeth Sherly, and John~P McCrae. 2020.
\newblock A survey of current datasets for code-switching research.
\newblock In \emph{Proceedings of ICACCS}.

\bibitem[{Joulin et~al.(2016)Joulin, Grave, Bojanowski, Douze, J{\'{e}}gou, and Mikolov}]{DBLP:journals/corr/JoulinGBDJM16}
Armand Joulin, Edouard Grave, Piotr Bojanowski, Matthijs Douze, Herv{\'{e}} J{\'{e}}gou, and Tom{\'{a}}s Mikolov. 2016.
\newblock Fasttext.zip: Compressing text classification models.
\newblock \emph{CoRR}, abs/1612.03651.

\bibitem[{Kakwani et~al.(2020)Kakwani, Kunchukuttan, Golla, Gokul, Bhattacharyya, Khapra, and Kumar}]{kakwani2020indicnlpsuite}
Divyanshu Kakwani, Anoop Kunchukuttan, Satish Golla, NC~Gokul, Avik Bhattacharyya, Mitesh~M Khapra, and Pratyush Kumar. 2020.
\newblock Indicnlpsuite: Monolingual corpora, evaluation benchmarks and pre-trained multilingual language models for indian languages.
\newblock In \emph{Findings of the ACL}.

\bibitem[{Khanuja et~al.(2021)Khanuja, Bansal, Mehtani, Khosla, Dey, Gopalan, Margam, Aggarwal, Nagipogu, Dave et~al.}]{khanuja2021muril}
Simran Khanuja, Diksha Bansal, Sarvesh Mehtani, Savya Khosla, Atreyee Dey, Balaji Gopalan, Dilip~Kumar Margam, Pooja Aggarwal, Rajiv~Teja Nagipogu, Shachi Dave, et~al. 2021.
\newblock Muril: Multilingual representations for indian languages.
\newblock \emph{arXiv preprint arXiv:2103.10730}.

\bibitem[{KhudaBukhsh et~al.(2020)KhudaBukhsh, Palakodety, and Carbonell}]{khudabukhsh2020harnessing}
Ashiqur~R KhudaBukhsh, Shriphani Palakodety, and Jaime~G Carbonell. 2020.
\newblock Harnessing code switching to transcend the linguistic barrier.
\newblock In \emph{Proceedings of IJCAI}.

\bibitem[{Kowsher et~al.(2022)Kowsher, Sami, Prottasha, Arefin, Dhar, and Koshiba}]{kowsher2022bangla}
M~Kowsher, Abdullah~As Sami, Nusrat~Jahan Prottasha, Mohammad~Shamsul Arefin, Pranab~Kumar Dhar, and Takeshi Koshiba. 2022.
\newblock Bangla-bert: transformer-based efficient model for transfer learning and language understanding.
\newblock \emph{IEEE Access}, 10:91855--91870.

\bibitem[{Krishnan et~al.(2022)Krishnan, Anastasopoulos, Purohit, and Rangwala}]{krishnan2022cross}
Jitin Krishnan, Antonios Anastasopoulos, Hemant Purohit, and Huzefa Rangwala. 2022.
\newblock Cross-lingual text classification of transliterated hindi and malayalam.
\newblock In \emph{Proceedings of Big Data}.

\bibitem[{Kumar et~al.(2020)Kumar, Ojha, Malmasi, and Zampieri}]{kumar2020evaluating}
Ritesh Kumar, Atul~Kr Ojha, Shervin Malmasi, and Marcos Zampieri. 2020.
\newblock Evaluating aggression identification in social media.
\newblock In \emph{Proceedings of TRAC}.

\bibitem[{Kumar et~al.(2018)Kumar, Reganti, Bhatia, and Maheshwari}]{KUMAR18.861}
Ritesh Kumar, Aishwarya~N. Reganti, Akshit Bhatia, and Tushar Maheshwari. 2018.
\newblock {Aggression-annotated Corpus of Hindi-English Code-mixed Data}.
\newblock In \emph{Proceedings of LREC}.

\bibitem[{Kumari and Singh(2020)}]{kumari2020ai_ml_nit_patna}
Kirti Kumari and Jyoti~Prakash Singh. 2020.
\newblock Ai\_ml\_nit\_patna@ trac-2: Deep learning approach for multi-lingual aggression identification.
\newblock In \emph{Proceedings of TRAC}.

\bibitem[{Kv{\aa}lseth(1989)}]{kvaalseth1989note}
Tarald~O Kv{\aa}lseth. 1989.
\newblock Note on cohen's kappa.
\newblock \emph{Psychological reports}, 65(1):223--226.

\bibitem[{Liu et~al.(2019)Liu, Ott, Goyal, Du, Joshi, Chen, Levy, Lewis, Zettlemoyer, and Stoyanov}]{liu2019roberta}
Yinhan Liu, Myle Ott, Naman Goyal, Jingfei Du, Mandar Joshi, Danqi Chen, Omer Levy, Mike Lewis, Luke Zettlemoyer, and Veselin Stoyanov. 2019.
\newblock Roberta: {A} robustly optimized {BERT} pretraining approach.
\newblock \emph{CoRR}, abs/1907.11692.

\bibitem[{Mahbub-ul Alam and Quyyum(2016)}]{mahbub2016sociolinguistic}
Ahmad Mahbub-ul Alam and Shaima Quyyum. 2016.
\newblock A sociolinguistic survey on code switching \& code mixing by the native speakers of bangladesh.
\newblock \emph{Journal of Manarat International University}, 6(1):8--9.

\bibitem[{Mazzocchi(2012)}]{langdetect}
Daniele Mazzocchi. 2012.
\newblock \href {https://github.com/Mimino666/langdetect} {langdetect: Language detection library}.
\newblock Python library.

\bibitem[{Mundra and Mittal(2022)}]{mundra2022fa}
Shikha Mundra and Namita Mittal. 2022.
\newblock Fa-net: fused attention-based network for hindi english code-mixed offensive text classification.
\newblock \emph{Social Network Analysis and Mining}, 12(1):100.

\bibitem[{Nayak and Joshi(2022)}]{nayak-joshi-2022-l3cube}
Ravindra Nayak and Raviraj Joshi. 2022.
\newblock {L}3{C}ube-{H}ing{C}orpus and {H}ing{BERT}: A code mixed {H}indi-{E}nglish dataset and {BERT} language models.
\newblock In \emph{Proceedings of WILDRE}.

\bibitem[{{Nick Doiron}(2023)}]{nick_doiron_2023}
{Nick Doiron}. 2023.
\newblock \href {https://doi.org/10.57967/hf/1066} {hindi-bert}.
\newblock Accessed: 2023-09-10.

\bibitem[{Nie(2023)}]{awesome_instruction_datasets}
Jianzhi Nie. 2023.
\newblock \href {https://github.com/jianzhnie/awesome-instruction-datasets} {Awesome instruction datasets}.
\newblock Accessed: 2023-09-10.

\bibitem[{OpenAI(2023)}]{openai2023gpt35turbo}
OpenAI. 2023.
\newblock \href {https://openai.com/blog/gpt-3-5-turbo-fine-tuning-and-api-updates} {Gpt-3.5 turbo fine-tuning and api updates}.
\newblock Accessed: 2023-08-28.

\bibitem[{Ranasinghe and Zampieri(2021)}]{ranasinghe2021evaluation}
Tharindu Ranasinghe and Marcos Zampieri. 2021.
\newblock An evaluation of multilingual offensive language identification methods for the languages of india.
\newblock \emph{Information}, 12(8):306.

\bibitem[{Rani et~al.(2020)Rani, Suryawanshi, Goswami, Chakravarthi, Fransen, and McCrae}]{rani2020comparative}
Priya Rani, Shardul Suryawanshi, Koustava Goswami, Bharathi~Raja Chakravarthi, Theodorus Fransen, and John~Philip McCrae. 2020.
\newblock A comparative study of different state-of-the-art hate speech detection methods in hindi-english code-mixed data.
\newblock In \emph{Proceedings of TRAC}.

\bibitem[{Ravikiran and Annamalai(2021)}]{ravikiran2021dosa}
Manikandan Ravikiran and Subbiah Annamalai. 2021.
\newblock Dosa: Dravidian code-mixed offensive span identification dataset.
\newblock In \emph{Proceedings of TRAC}.

\bibitem[{Rosenthal et~al.(2021)Rosenthal, Atanasova, Karadzhov, Zampieri, and Nakov}]{rosenthal-etal-2021-solid}
Sara Rosenthal, Pepa Atanasova, Georgi Karadzhov, Marcos Zampieri, and Preslav Nakov. 2021.
\newblock {SOLID}: A large-scale semi-supervised dataset for offensive language identification.
\newblock In \emph{Findings of the ACL}.

\bibitem[{Sai and Sharma(2020)}]{sai2020siva}
Siva Sai and Yashvardhan Sharma. 2020.
\newblock Siva@ hasoc-dravidian-codemix-fire-2020: Multilingual offensive speech detection in code-mixed and romanized text.
\newblock In \emph{Proceedings of FIRE}.

\bibitem[{Sanh et~al.(2019)Sanh, Debut, Chaumond, and Wolf}]{DBLP:journals/corr/abs-1910-01108}
Victor Sanh, Lysandre Debut, Julien Chaumond, and Thomas Wolf. 2019.
\newblock Distilbert, a distilled version of {BERT:} smaller, faster, cheaper and lighter.
\newblock In \emph{Proceedings of EMC2}.

\bibitem[{Santy et~al.(2021)Santy, Srinivasan, and Choudhury}]{santy2021bertologicomix}
Sebastin Santy, Anirudh Srinivasan, and Monojit Choudhury. 2021.
\newblock Bertologicomix: How does code-mixing interact with multilingual bert?
\newblock In \emph{Proceedings of AdaptNLP}.

\bibitem[{Sarkar et~al.(2021)Sarkar, Zampieri, Ranasinghe, and Ororbia}]{sarkar-etal-2021-fbert-neural}
Diptanu Sarkar, Marcos Zampieri, Tharindu Ranasinghe, and Alexander Ororbia. 2021.
\newblock f{BERT}: A neural transformer for identifying offensive content.
\newblock In \emph{Findings of the ACL}.

\bibitem[{Singh(1985)}]{singh1985grammatical}
Rajendra Singh. 1985.
\newblock Grammatical constraints on code-mixing: Evidence from hindi-english.
\newblock \emph{Canadian Journal of Linguistics/Revue canadienne de linguistique}, 30(1):33--45.

\bibitem[{Sreelakshmi et~al.(2020)Sreelakshmi, Premjith, and Soman}]{sreelakshmi2020detection}
K~Sreelakshmi, B~Premjith, and KP~Soman. 2020.
\newblock Detection of hate speech text in hindi-english code-mixed data.
\newblock \emph{Procedia Computer Science}, 171:737--744.

\bibitem[{Thara and Poornachandran(2018)}]{thara2018code}
S~Thara and Prabaharan Poornachandran. 2018.
\newblock Code-mixing: A brief survey.
\newblock In \emph{Proceedings of ICACCI}.

\bibitem[{Vasantharajan and Thayasivam(2021)}]{vasantharajan2021hypers}
Charangan Vasantharajan and Uthayasanker Thayasivam. 2021.
\newblock Hypers@ dravidianlangtech-eacl2021: Offensive language identification in dravidian code-mixed youtube comments and posts.
\newblock In \emph{Proceedings of DravidianLangTech}.

\bibitem[{Wadud et~al.(2021)Wadud, Hamid, Monowar, and Alamri}]{mridha2021boost}
Md~Anwar~Hussen Wadud, Md~Abdul Hamid, Muhammad~Mostafa Monowar, and Atif Alamri. 2021.
\newblock L-boost: Identifying offensive texts from social media post in bengali.
\newblock \emph{Ieee Access}, 9:164681--164699.

\bibitem[{Zampieri et~al.(2019)Zampieri, Malmasi, Nakov, Rosenthal, Farra, and Kumar}]{zampieri-etal-2019-semeval}
Marcos Zampieri, Shervin Malmasi, Preslav Nakov, Sara Rosenthal, Noura Farra, and Ritesh Kumar. 2019.
\newblock {S}em{E}val-2019 task 6: Identifying and categorizing offensive language in social media ({O}ffens{E}val).
\newblock In \emph{Proceedings of SemEval}.

\end{thebibliography}
\bibliographystyle{acl_natbib}

\appendix

\section{Examples of Misclassified Instances}
\label{sec:appendix_a}
\includegraphics[width = \linewidth]{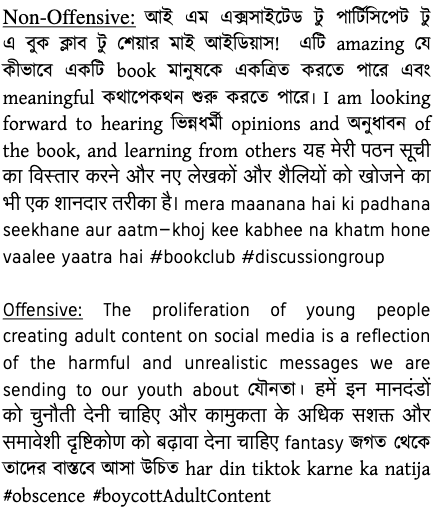}

\end{document}